\documentclass[sigconf,screen]{acmart}

\def\Snospace~{\S{}}

\copyrightyear{2023}
\acmYear{2023}
\setcopyright{rightsretained} 
\acmConference[HRI '23]{Proceedings of the 2023 ACM/IEEE International Conference on Human-Robot Interaction}{March 13--16, 2023}{Stockholm, Sweden}
\acmBooktitle{Proceedings of the 2023 ACM/IEEE International Conference on Human-Robot Interaction (HRI '23), March 13--16, 2023, Stockholm, Sweden}
\acmDOI{10.1145/3568162.3578623}
\acmISBN{978-1-4503-9964-7/23/03}

\begin{document}

\title[Online Language Corrections via Shared Autonomy]{\textit{``No, to the Right''} -- Online Language Corrections for Robotic Manipulation via Shared Autonomy}

\author{Yuchen Cui}
\authornote{Both authors contributed equally to this research.}
\email{yuchenc@cs.stanford.edu}
\affiliation{
    \institution{Stanford University}
    \city{Stanford}
    \state{CA}
    \country{USA}
}

\author{Siddharth Karamcheti}
\authornotemark[1]
\email{skaramcheti@cs.stanford.edu}
\affiliation{
    \institution{Stanford University}
    \city{Stanford}
    \state{CA}
    \country{USA}
}

\author{Raj Palleti}
\affiliation{
    \institution{Stanford University}
    \city{Stanford}
    \state{CA}
    \country{USA}
}

\author{Nidhya Shivakumar}
\affiliation{
    \institution{The Harker School}
    \city{San Jose}
    \state{CA}
    \country{USA}
}

\author{Percy Liang}
\affiliation{
    \institution{Stanford University}
    \city{Stanford}
    \state{CA}
    \country{USA}
}

\author{Dorsa Sadigh}
\affiliation{
    \institution{Stanford University}
    \city{Stanford}
    \state{CA}
    \country{USA}
    \vspace{3mm}
}

\renewcommand{\shortauthors}{Cui \& Karamcheti et. al.}

\begin{abstract}
Systems for language-guided human-robot interaction must satisfy two key desiderata for broad adoption: \textit{adaptivity} and \textit{learning efficiency}. Unfortunately, existing instruction-following agents cannot adapt, lacking the ability to incorporate online natural language supervision, and even if they could, require hundreds of demonstrations to learn even simple policies. In this work, we address these problems by presenting Language-Informed Latent Actions with Corrections (LILAC), a framework for incorporating and adapting to natural language corrections -- ``to the right'', or ``no, towards the book'' -- \textit{online, during execution}. We explore rich manipulation domains within a \textit{shared autonomy} paradigm. Instead of discrete turn-taking between a human and robot, LILAC \textit{splits agency} between the human and robot: language is an input to a learned model that produces a meaningful, low-dimensional control space that the human can use to guide the robot. Each real-time correction refines the human's control space, enabling precise, extended behaviors -- with the added benefit of requiring only a handful of demonstrations to learn. We evaluate our approach via a user study where users work with a Franka Emika Panda manipulator to complete complex manipulation tasks. Compared to existing learned baselines covering both open-loop instruction following and single-turn shared autonomy, we show that our corrections-aware approach obtains higher task completion rates, and is subjectively preferred by users because of its reliability, precision, and ease of use.\footnote{
    Project website with videos \& study interface: \url{https://sites.google.com/view/hri-lilac}. \\
    \hbox{\hspace{1pt}} Code for data collection, model definition, training, and evaluation: \\
    \hbox{\hspace{1mm} \url{https://github.com/Stanford-ILIAD/lilac}}.
}
    
\end{abstract}

\ccsdesc[500]{Computing methodologies~Cooperation and coordination}
\ccsdesc[300]{Computing methodologies~Natural language processing}
\ccsdesc[300]{Computing methodologies~Learning from demonstrations}

\keywords{Online corrections, language \& shared autonomy, robot learning}
\maketitle

\section{Introduction}
\label{sec:introduction}
\begin{figure}[t!]
    \centering
    \includegraphics[width=\columnwidth]{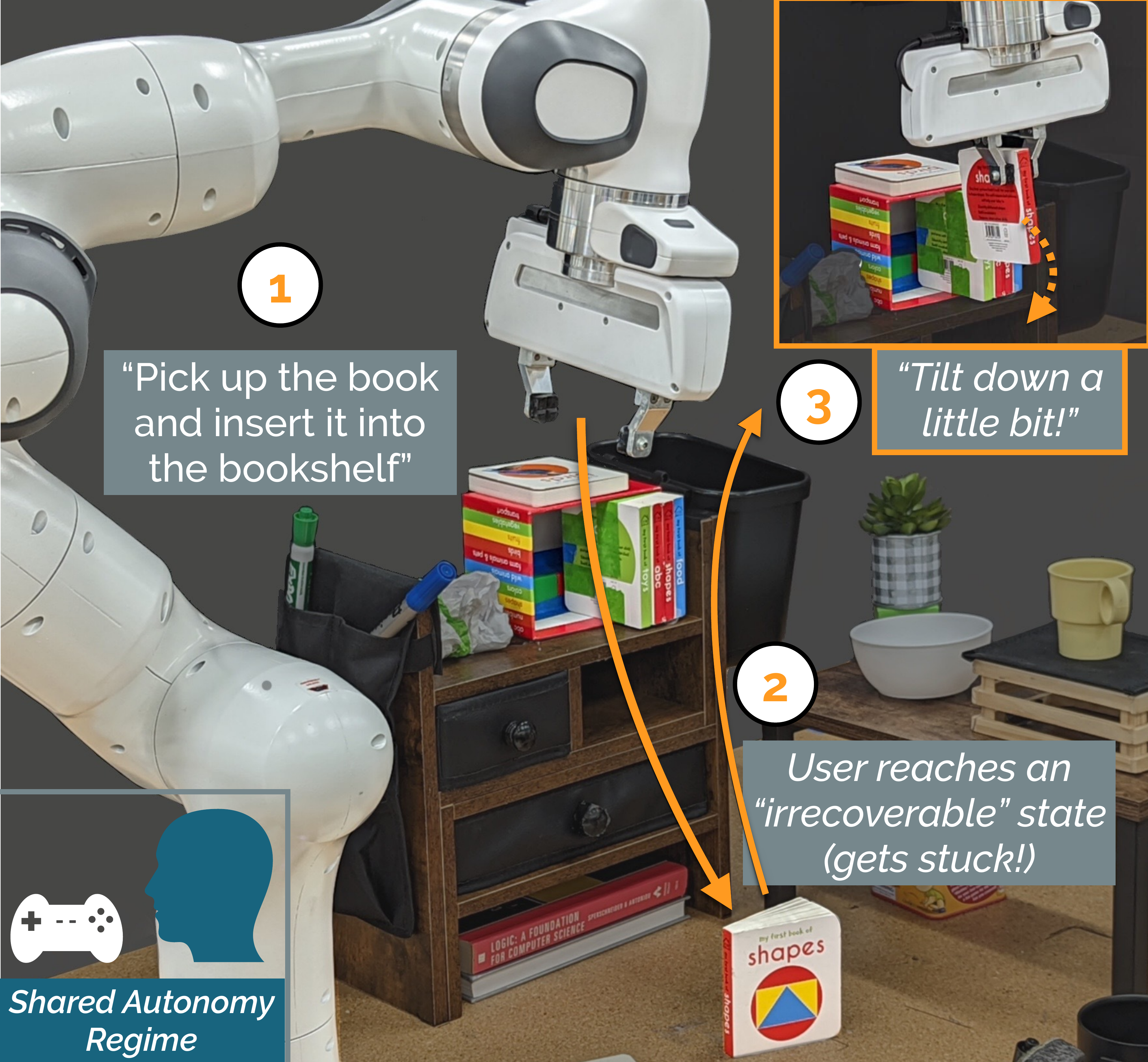}
    \vspace*{-7mm}
    \caption{LILAC: Whereas prior work only allows for issuing a \textit{single} language utterance for the entire task (``Pick up the book and insert it into the bookshelf'' -- solid line), our approach allows users to provide \textit{language corrections} at any point during execution, allowing the robot to adapt online (``Tilt down a little bit!'' -- right window).}
    \label{fig:front-figure}
\end{figure}

\begin{figure*}[t!]
    \centering
    \includegraphics[width=0.95\linewidth]{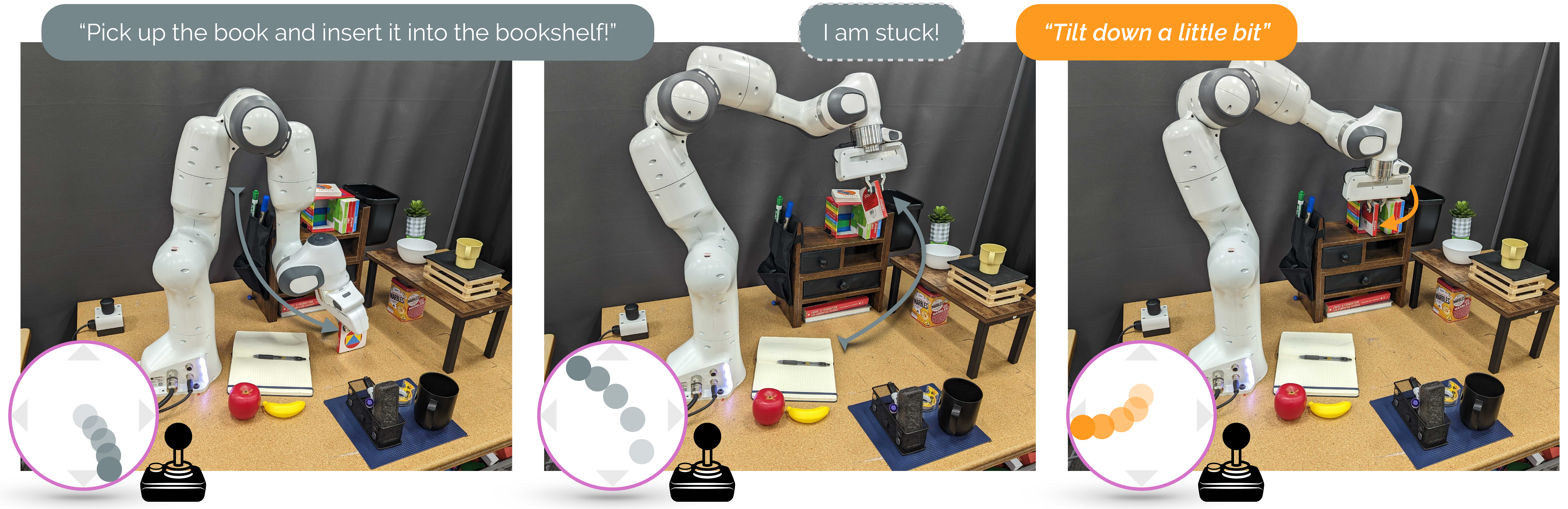}
    \vspace*{-3mm}
    \caption{A user interacts with our system. [Left] The user utters ``pick up the book and insert it into the bookshelf,'' inducing a low-dimensional controller (depicted with the joystick and shaded inputs). [Middle] This control space is state and language-conditioned: pressing down brings the end-effector close to the book, while holding up/left after grasping the book moves the end-effector towards the shelf. However, this \textit{static} controller is not enough; the user gets stuck! [Right] Our approach allows users to provide real-time corrections (``tilt down a little bit'') refining the control space so the user can complete the task.}
    \label{fig:motivating-example}
\end{figure*}

Research in natural language for robotics has focused on \textit{dyadic, turn-based interactions} between humans and robots, often in the instruction following regime \citep{tellex2011understanding, artzi2013weakly, thomason2015learning, arumugam2017accurately}. In this paradigm a human gives an instruction, \textit{then} the robot executes autonomously -- simultaneously resolving the human's goal as well as planning a course of actions to execute in the environment, without any additional user input. This explicit division of agency between humans and robots places a tremendous burden on learning; existing systems either require large amounts of language-aligned demonstration data to learn policies \citep{chevalierboisvert2019babyai,stepputtis2020lcil,lynch2020grounding,shridhar2021clipport}, or make other restrictive assumptions about known environment dynamics, in addition to the ability to perform perfect object localization and affordance prediction to plug into task and motion planners \citep{matuszek2012grounded, kollar2013towards}.

Coupled with the sample inefficiency of these approaches is their \textit{lack of adaptivity}. Consider the robot in \autoref{fig:front-figure}, trying to execute ``Pick up the book and insert it into the bookshelf.'' This is a long-horizon task with several critical states requiring precise manipulation -- from grasping the book by its spine, to raising it above the table without hitting the side table, to lining it up precisely with the bookshelf for insertion (with less than a few millimeters of clearance on either side). In such circumstances, even the best existing approaches fail to complete the task repeatably. Yet these ``task failures'' are often predictable and recoverable. A user watching a robot diving towards a glass bowl knows that catastrophe is seconds away, and how to avert it; similarly, fine-grained errors such as a missed grasp, or a misaligned end-effector are similarly fixable -- as long as the user is provided with the right mechanism to adapt the robot's behavior.

One way to enable such adaptation is through \textit{natural language corrections} -- from the simple ``left!'' or ``tilt down a little bit'' (as in \autoref{fig:front-figure}), to the more complex ``no, towards the \textit{blue} marker.'' While recent work tries to get at the spirit of this idea by learning from dialogue \citep{thomason2019visdial, thomason2019improving}, post-hoc corrections \citep{coreyes2019guiding, sharma2022correcting, bucker2022reshaping, bucker2022latte}, or implicit feedback \citep{karamcheti2020decomposition}, none of these approaches work in \textit{real-time}. Instead, we argue that scalable systems for language-driven human-robot interaction must be able to handle online corrections in a manner that is both \textit{adaptive} and \textit{sample efficient}. 

We introduce a novel approach -- \textbf{LILAC: Language-Informed Latent Actions with Corrections} -- that presents a generalizable framework for adapting to \textit{online} natural language corrections built within a \textit{shared autonomy} \citep{dragan2013policy, argall2018autonomy, javdani2018shared, losey2021latentactions} paradigm for human-robot collaboration. With LILAC, a user provides a stream of language utterances, starting with a high-level goal (``Pick up the book and insert it into the bookshelf''), with each utterance shaping the control the user is afforded over the robot. At any point during execution, a user can provide a new utterance -- a correction like ``tilt down a little bit!'' -- which updates that control space \textit{online}, reflecting the user's intent in real time. Working in a shared autonomy setting like this gives us the \textit{adaptivity} mentioned above, but also allows us to develop correction-aware systems with extreme \textit{sample efficiency}. With LILAC, we can learn to perform complex manipulation tasks like those in \autoref{fig:front-figure} from 10-20 demonstrations instead of the thousands to tens of thousands of demonstrations required by fully autonomous imitation or reinforcement learning approaches\citep{chevalierboisvert2019babyai, luketina2019survey, lynch2020grounding, jang2021bcz}. These gains are rooted in the idea of \textit{splitting agency} between the human and robot; during execution, both parties influence the ultimate actions of the robot, sharing the burden of reasoning over actions.

We evaluate LILAC via a within-subjects user study ($n=12$), where users complete a complex set of manipulation tasks on a Franka Emika Panda arm using LILAC and two baselines -- the state-of-the-art language-informed latent actions (LILA) model \citep{karamcheti2021lila}, as well as a fully autonomous language-conditioned imitation learning approach. We find that LILAC obtains higher task success rates than either baseline because of its ability to adapt given online language instructions, and that users qualitatively find LILAC to be more reliable, precise, and easy to use.

\section{Motivating Example}
\label{sec:motivating-example}
The learned latent actions paradigm \citep{losey2020latent, jeon2020sharedlatent} was initially conceived of in the context of assistive teleoperation; given users with limited mobility, finding an intuitive manner of controlling a 7-DoF+ assistive robotic arm with low-dimensional controllers (e.g., a 2-DoF joystick attached to a wheelchair) is extremely difficult. Naive approaches for mapping the high-dimensional robotic control problem to a low-DoF interface -- for example, by controlling the (x/y, z/roll, pitch/yaw) axes of the end-effector independently with the joystick -- lead to high amounts of user discomfort, with frequent mode-switching, imprecise controls, and high cognitive load for users \citep{herlant2016assistive, argall2018autonomy}. Learned latent actions -- and specifically, the latest work on Language-Informed Latent Actions (LILA) \cite{karamcheti2021lila} -- offer a compromise: use a small number of task-specific demonstrations to learn a nonlinear mapping from joystick axes to end-effector control axes, such that each axis of the joystick represents semantically meaningful movement through task space. 

As a concrete example, consider \autoref{fig:motivating-example} for the task of ``pick up the book and insert it into the bookshelf.'' LILA learns a single, \textit{static} mapping to use for the entirety of the episode, and hits a key failure mode; due to compounding errors as the user navigated the book from the table up towards the shelf, the end-effector is misaligned with the shelf, making a clean insertion impossible! This is where we need \textit{online language corrections} -- the mechanism that allows the user to quickly diagnose the problem and refine the robot's behavior. With LILAC, the user provides the correction ``tilt down a little bit,'' in the midst of execution and switch into a new control space. Pressing left on the joystick now provides explicit, precise control over the robot's orientation, allowing the user to even-out the end-effector and complete the task.

\section{Related Work}
\label{sec:related-work}
\begin{figure*}[t!]
    \centering
    \includegraphics[width=0.95\linewidth]{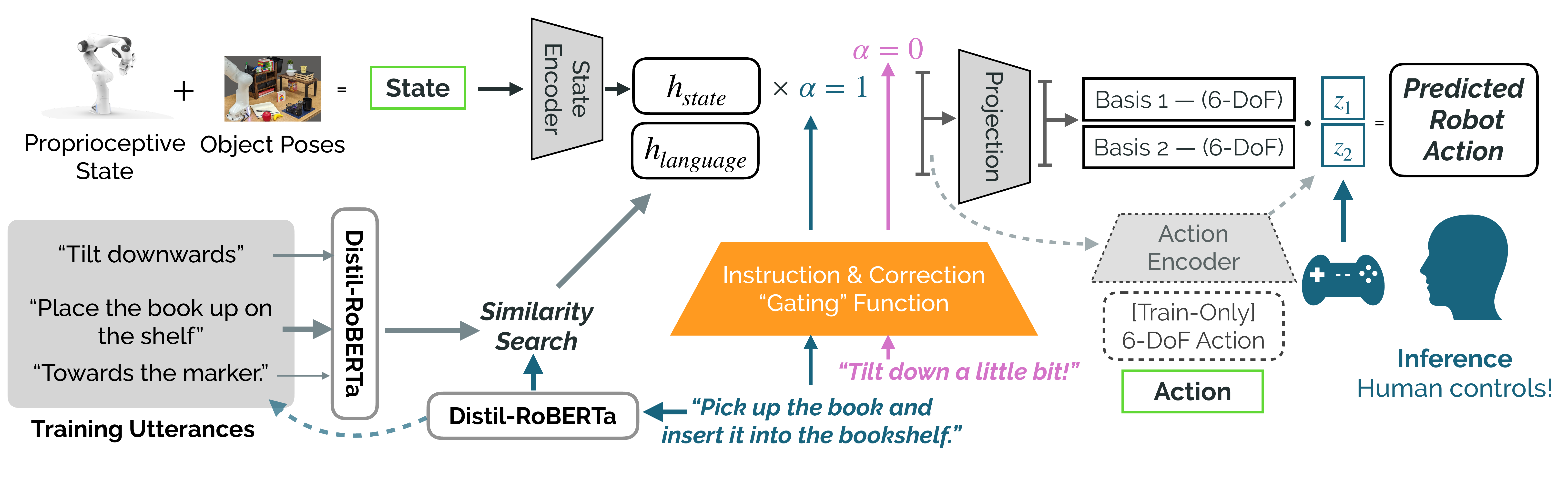}
    \vspace*{-3mm}
    \caption{LILAC Overview -- solid lines represent the inference pipeline, while dashed lines indicate training-only steps. Part of LILAC's ability to incorporate language corrections efficiently is the ``gating'' module (orange) which controls the amount of state-context for a given input -- for example, grounding a correction such as ``tilt down a little bit'' requires no state context ($\alpha = 0$), whereas a high-level instruction such as ``pick up the book and insert it into the bookshelf'' does require context ($\alpha = 1$). We use GPT-3, a  pretrained language model, to provide $\alpha$ (see \autoref{subsec:instructions-vs-corrections} for discussion).}
    \label{fig:system}
\end{figure*}

LILAC builds off of a rich body of work spanning methods for incorporating language corrections, learning language-conditioned policies, and incorporating other forms of corrective feedback.

\smallskip

\noindent \textbf{Incorporating Language Corrections for Manipulation.} Most relevant to our approach are recent methods for incorporating various types of natural language corrections in the context of robotic manipulation. These methods can be stratified based on the assumptions they make, and \textit{when} during execution a user provides a correction. For example, \citet{broad2017realtime} enables data-efficient online corrections (similar to LILAC) using distributed correspondence graphs to ground language, via use of a semantic parser that maps language to a predefined space of correction groundings; these groundings are brittle and hand-designed, additionally requiring access to a motion planner (and fully known environment dynamics) to identify a fulfilling set of actions for the robot to execute. In contrast, later work in incorporating corrections removes the need for brittle, hand-designed correction primitives, instead using \textit{post-hoc} corrections provided at the end of task execution to define composable cost functions that are fed to a trajectory optimizer \citep{sharma2022correcting}; the post-hoc nature of these corrections is limiting, especially in cases where tasks have irreversible or ``hard-to-reset'' components, and the trajectory optimizer requires non-trivial knowledge of the environment. Later work relaxes these prior knowledge assumptions, but can only incorporate correction information post-hoc, directly ``modifying'' trajectory waypoints following a language correction in both 2D \citep{bucker2022reshaping} and 3D \citep{bucker2022latte} environments using massive datasets of paired corrections and demonstrations. In contrast to these approaches, LILAC is a shared autonomy approach that operates \textit{online, in real-time}, without a need for massive amounts of data, prior environment dynamics, or full state knowledge.

\medskip

\noindent \textbf{Learning Language-Conditioned Policies.}  More general than incorporating corrections is a tremendous body of work on learning language-conditioned policies in both the full and shared autonomy regimes. Early work in this space used semantic parsers to map natural language instructions to predefined motion planning primitives, given modest sized datasets \citep{tellex2011understanding, kollar2013towards, artzi2013weakly, arumugam2017accurately}. While these approaches were able to accomplish a limited range of tasks with high reliability, reliance on predefined primitives and motion planners made it hard to scale these approaches to more complex manipulation domains, where environment knowledge is hard to come by, and hand-defined primitives are brittle and limiting. As a result, more recent work in this space learn language-conditioned policies directly via imitation learning from large datasets of paired (language, demonstration) pairs \citep{lynch2020grounding, stepputtis2020lcil, jang2021bcz, shridhar2021clipport, mees2022matters}. While expressive, these approaches are still tremendously data hungry, requiring hundreds or thousands of examples to learn even the simplest tasks. To address the sample efficiency problem, other work such as LILA \citep{karamcheti2021lila} have turned to the shared autonomy regime, learning collaborative human-robot policies from orders of magnitude fewer demonstrations. LILA is the starting point of our proposed approach.

\medskip

\noindent \textbf{Incorporating Other Forms of Corrective Feedback.} Other approaches tackle learning from other forms of corrective feedback, such as physical corrections \citep{losey2018review, li2021learning}, targeted interventions wherein a human fully assumes control over a robot via remote teleoperation \citep{kelly2019hgdagger,mandlekar2020human,hoque2021thriftydagger}, critiques \citep{chernova2014robot, cui2018active}, as well as trajectory preferences \citep{christiano2017deep, biyik2018batch}. More recently, \citet{schmittle2020learning} have proposed a meta-algorithm for online learning from multiple types of corrective feedback (excluding language). While this work is promising, we focus our approach on language corrections, a natural communication modality for human users. Learning to incorporate language corrections also allows for \textit{transfer across tasks} as correction language such as ``to the left'' are general and often independent of the current state of the robot, whereas other correction modalities can be more context-dependent.

\section{LILAC: Framing Corrections}
\label{sec:lilac}
LILAC builds off of LILA as introduced by \citet{karamcheti2021lila} by incorporating natural language corrections during the course of execution. The LILAC architecture is depicted in \autoref{fig:system}; solid lines denote inference, while dashes denote training. LILAC incorporates natural language corrections in a data-driven way; this work focuses on directional corrections (e.g., ``to the left'') and referential corrections (e.g., ``towards the blue marker'') -- all with a generalizable and scalable procedure that can be extended to other more complex types of corrections. This section outlines all the elements of our approach.

\subsection{Problem Statement}
\label{subsec:lilac-problem-statement}

Our setting is that of a sequential decision making problem defined by elements $(\mathcal{S}, \mathcal{A}, \mathcal{T}, \mathcal{U}, \mathcal{C^*}, \mathcal{Z})$ where $s \in \mathcal{S} \subseteq \mathbb{R}^n$ denotes the state of the robot and environment, $a \in \mathcal{A} \subseteq \mathbb{R}^k$ denotes a robot's $k$-dimensional action (in our case, a 6-DoF delta in end-effector pose -- Cartesian coordinates for position and Euler angles for orientation), and $\mathcal{T}: \mathcal{S} \times \mathcal{A} \rightarrow \mathcal{S}$ is a (stochastic) unobserved transition function. Furthermore, $u \in \mathcal{U}$ denotes a high-level natural language instruction provided by the user, $\mathbf{c} \in \mathcal{C^*}$ denotes the ordered (possibly empty) stack of natural language corrections the user has provided, and $z \in \mathcal{Z} \subseteq \mathbb{R}^d$ where $d \ll k$ denotes a user-provided input via their low-dimensional control device (e.g., a 2-DoF joystick). Users can provide an arbitrary number of online corrections \textit{throughout} the episode to adapt the robot's behavior.

The goal of LILAC is to learn a function $\mathcal{F}_\theta(s_t, z_t, u_t, \mathbf{c}_t): \mathcal{S} \times \mathcal{Z} \times \mathcal{U} \times \mathcal{C^*} \rightarrow \mathcal{A}$ that maps the current robot and environment state $s_t$, low-dimensional control input $z_t$, initial high-level utterance $u$ provided by the user (held constant throughout the given episode), and (possibly empty) stack of language corrections $\mathbf{c}_t$ to a high-dimensional robot action $a_t$ that is to be executed in the environment. The corresponding low-DoF control manifold $\bigcup_{z_t \in \mathcal{Z}} \mathcal{F}_\theta(s_t, z_t, u_t, \mathbf{c}_t)$ provides an intuitive interface for the user to maneuver the robot towards satisfying the task in question. At each new timestep $t+1$, a user can either provide a new language correction $c'$ which is ``pushed'' onto the stack, press a button to ``pop'' their latest correction off of the stack $\mathbf{c}_t$ signalling that their correction has been addressed, or provide a control input $z_t$ that is mapped to the corresponding robot action $a_t$.

\subsection{Modeling: Inference \& Learning}
\label{subsec:lilac-learning}

Given the current state $s$ and language $u$ and $\mathbf{c}$, $\mathcal{F}_\theta$ maps low-dimensional user control inputs $z$ to the high-dimensional robot actions $a$. Crisply, we define $\hat{a} = \mathcal{F}_\theta(s, z, u, \mathbf{c})$ as:
\begin{align*}
    h_\text{state} \in \mathbb{R}^m &= \texttt{EncodeState}_\theta(s) \\
    h_\text{language} \in \mathbb{R}^m &= \texttt{EncodeLanguage}_\theta(u, \mathbf{c}) \\
    \alpha \in [0, 1] &= \texttt{GPTGating}(u, \mathbf{c}) \\
    h_\text{gated} \in \mathbb{R}^m &= \alpha \cdot h_\text{state} + (1 - \alpha) \cdot \text{bias}_\theta \\
    h_\text{fused} \in \mathbb{R}^m &= \texttt{FiLM}_\theta(h_\text{gated}, h_\text{language}) \\ 
    B_\text{bases} \in \mathbb{R}^{k \times d} &= \text{Gram-Schmidt}(\texttt{Projection}_\theta(h_\text{fused})) \\
    \hat{a} \in \mathbb{R}^k &= B_\text{bases} \cdot z
\end{align*}
where $m$ is a hyperparameter denoting the hidden dimensionality of the model (we set $m=128$ for this work). 

We first learn a state encoder $\texttt{EncodeState}_\theta$. The state space we use in this work consists of the robot's proprioceptive state (joint angles \& end-effector pose) concatenated with a vector of $(x, y, z)$ positions for each object in the scene; $\texttt{EncodeState}_\theta$ is a two-layer MLP that takes this input and outputs $h_\text{state}$. 

To encode language ($u$ and $\mathbf{c}$), we adopt a last-in-first-out strategy for selectively encoding utterances, only encoding on the most recent utterance -- $u$ at the beginning of an interaction, then the most recent correction $c'$. We embed this utterance with a frozen variant of the Distil-RoBERTa language model \citep{sanh2019distilbert} as released by the Sentence-BERT project \citep{reimers2019sentence}. This process is backed by an ``unnatural language processing'' nearest neighbors index \citep{marzoev2020unnatural} where inference-time utterances are mapped onto the closest existing training exemplars, which are then retrieved and fed to the rest of the model. This process, similar to that used in LILA \citep{karamcheti2021lila} prevents LILAC from degenerating in the presence of slight variations of language, which could lead to practical user safety issues. $\texttt{EncodeLanguage}_\theta$ is another two-layer MLP that takes in the retrieved embedding and outputs $h_\text{language}$.

Next, we consider how to fuse the state and language embeddings. A key component of LILAC and its ability to remain data efficient is the GPT-3 \citep{brown2020gpt3} ``gating'' module (shown in orange in \autoref{fig:system}), and denoted by the scalar value $\alpha \in [0, 1]$ in the equations above. The gating value $\alpha$ reflects a simple insight -- certain utterances require different amounts of object/state dependence. For example, an utterance such as ``go left'' does not require reasoning over any objects in the environment $(\alpha = 0)$. A detailed discussion on gating and how we operationalize GPT-3 can be found in \autoref{subsec:instructions-vs-corrections}. We use this gating value $\alpha$ to module the amount of state information in $h_\text{gated}$ by taking the convex combination of $\alpha$ with $h_\text{state}$ and a vector $\text{bias}_\theta$, where $\alpha = 0$ obviates $x_\text{state}$. 

We incorporate language by using FiLM \citep{perez2018film}, mapping $h_\text{language}$ to affine transformation parameters $\gamma \in \mathbb{R}^m$ and $\beta \in \mathbb{R}^m$ via separate learned two-layer MLPs. We then apply the resulting transformation elementwise to $h_\text{gated}$, producing $h_\text{fused} = \gamma \cdot h_\text{gated} + \beta$.

Finally, we predict basis vectors $B_\text{bases} \in \mathbb{R}^{k \times d}$ (recall that $k$ is the dimensionality of the high-DoF robot action space, while $d$ is the dimensionality of the user's low-DoF control interface). $B_\text{bases}$ uniquely defines the user's control manifold for the given timestep; we learn a two-layer MLP $\texttt{Projection}_\theta$ that takes $h_\text{fused}$ and outputs a matrix with the required dimensions. To serve as an appropriately conditioned control manifold, we run modified Gram-Schmidt to orthonormalize the bases. The final high-DoF robot action $\hat{a} \in \mathbb{R}^k$ is then the matrix-vector product between $B_\text{bases}$ and the user's input $z \in \mathbb{R}^d$.

\medskip 

\noindent \textbf{Learning from Language \& Demonstrations.} To learn $\mathcal{F}_\theta$, we assume a dataset of ($u$ = language, $\tau$ = trajectory) pairs, where each trajectory is comprised of a sequence of $(s, a)$ pairs; $\tau = \{(s_1, a_1) \ldots (s_T, a_T)\}$, and a desired latent action dimensionality $d$ (e.g., $d = 2$ in this work for the number of axes on a joystick). The action space we use are deltas in end-effector space (Cartesian position, Euler angle orientation). Note that these inputs don't fully line up with the signature of $\mathcal{F}_\theta$ -- notably, \textit{we do not} have access to ``ground-truth'' latent actions $z$ for each given robot action $a$. To address this, we adopt the insight used in prior work on learned latent actions \citep{losey2020latent, karamcheti2021vla, karamcheti2021lila}: frame the training process as learning a state-and-language conditional autoencoder, using \textit{compression} as a way to induce meaningful latent action control manifolds. 

Specifically, we implement this by adding a layer to compress high-DoF robot actions down to a $d$-dimensional latent:
\begin{align*}
    z_\text{compressed} \in \mathbb{R}^d &= \texttt{Compress}_\theta(a) \\
    a_\text{reconstruct} \in \mathbb{R}^k &= B_\text{bases} \cdot z_\text{compressed}
\end{align*}
where $B_\text{bases}$ is computed from state and language as above.

Given this reconstruction objective, we can write a compact loss function for training:  $\mathcal{L}(\theta) = ||a - a_\text{reconstruct}||_2^2$ -- in other words, minimize the mean squared error between the action high-DoF robot action $a$ and $a_\text{reconstruct}$. $\texttt{Compress}_\theta$ is implemented as a two-layer MLP, and is discarded after training.

\subsection{Gating Instructions vs. Corrections}
\label{subsec:instructions-vs-corrections}

Key to scaling LILAC is the insight that various forms of correction language are generalizable \textit{across states} -- in other words, different language utterances require different amounts of object/environment state-dependence. Formally defining the ``state-dependence'' of a language utterance is hard; one heuristic might be to categorize different utterances based on the number of \textit{referents} present; an utterance like ``grab the thing on the side table and place it on the table'' as in \autoref{fig:front-figure} has 3 referents, indicating a large degree of state dependence; the robot \textit{must} ground the utterance in the objects of the environment to resolve the correct behavior. However, an utterance like ``no, to the left!'' has no explicit referents; one can resolve the utterance by relying solely on the user's static reference frame and induced deltas in end-effector space.

To operationalize this idea with LILAC, further contributing to the sample efficiency of our approach,\footnote{A valid question is why not treat all utterances as requiring uniform, or the same amount of state-dependence; the answer is rooted in the small data regime we operate in. We'd need to collect several instances of the same correction ``to the left'' in different states to generalize, whereas with LILAC ``gating'' approach, we only need one!} we use a \textit{gating} function (orange, in \autoref{fig:system}) that given language, predicts a discrete value $\alpha \in \{0, 1\}$. A value of 0 signifies a state-independent utterance -- for example, the correction ``tilt down a little bit.'' Appropriately, in our architecture, this zeroes out any state-dependent information (see the $\alpha$ term in \autoref{fig:system}), and predicts an action solely based on the provided language. Critically, the fused representation of language and state provided to the rest of the network defining in $\mathcal{F}_\theta$ is modulated by $\alpha$ -- more detail in \autoref{subsec:reproducibility}.

\medskip

\noindent \textbf{Using GPT-3 to Identify Corrections}. In this work, we construct a prompt harness with GPT-3 \citep{brown2020gpt3} to output $\alpha$. We do this because characterizing the state-dependence of an utterance is difficult; while the aforementioned reference counting heuristic may work in some cases, utterances such as ``no, the blue!'' have implicit referents (in this case, perhaps a marker, or cup) that \textit{need to be grounded in the environment state}. Many other phenomena make it hard to define heuristics for computing $\alpha$ -- anaphora, null referents (``move the robot left''), etc. Instead of crafting grammars or heuristics, we opt to tap into the power of large language models with \textit{in-context learning} abilities, that learn to extrapolate given a prompt and small set of examples. We specifically build off of GPT-3 \texttt{text-davinci-002} (175B parameters) \citep{brown2020gpt3}. To define our prompt, we allocated a 10 minute budget to iteratively engineer the input/output examples and task description, using a held-out set of 5 language utterances to provide signal.\footnote{Prompt: \url{https://github.com/Stanford-ILIAD/lilac/tree/main/scripts/alphas.py}.}

\subsection{Reproducibility}
\label{subsec:reproducibility}

\begin{figure*}[t!]
    \centering
    \includegraphics[width=0.95\linewidth]{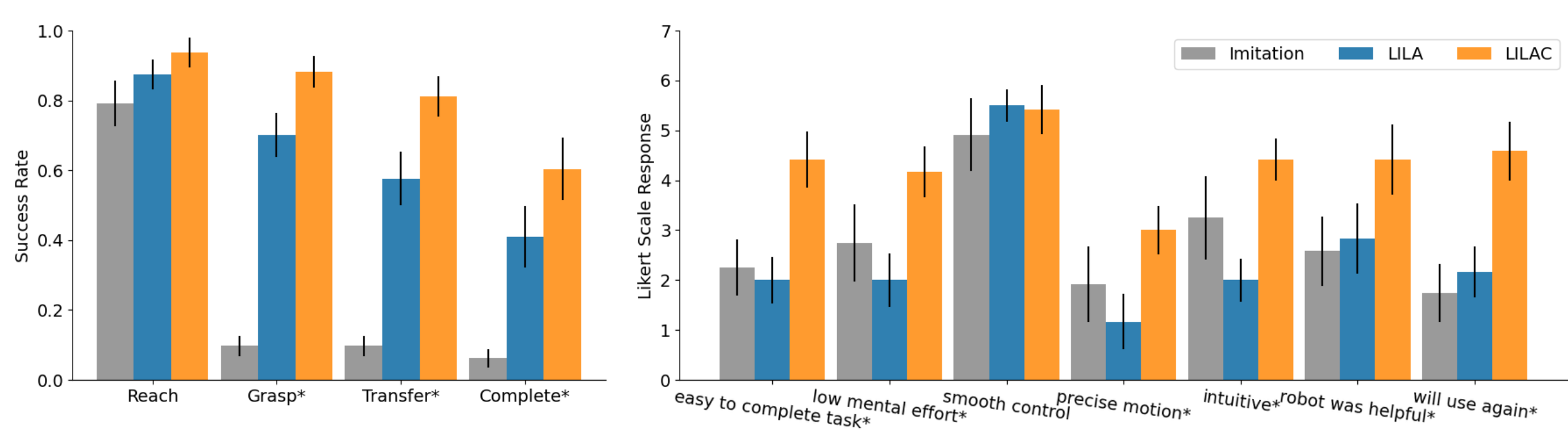}
    \vspace*{-3mm}
    \caption{Results from our user study ($n = 12$) across three conditions: 1) Language-Conditioned Imitation Learning, 2) Language-Informed Latent Actions (LILA) -- an instantiation of language-informed shared autonomy \textit{without} online corrections, and 3) LILAC -- our approach where users can provide online corrections at any point during robot execution.}
    \label{fig:result-plots}
\end{figure*}

To facilitate reproducibility and future work, we release an open-source codebase (\url{https://github.com/Stanford-ILIAD/lilac}) with the complete pipeline spanning data collection, model definition, training, and real-robot deployment.

\medskip

\noindent \textbf{Model Architecture.} All MLPs detailed in \autoref{subsec:lilac-learning} use $m=128$ and the \texttt{GELU} activation \citep{hendrycks2016gelu}. For stability, we add a single Batch Normalization layer \citep{ioffe2015batch} before feeding the concatenated state representation to the state encoder. As implemented, LILAC is extremely lightweight at only 188K parameters.

\medskip

\noindent \textbf{Training Details.} Training LILAC is efficient, and can be run on consumer laptop CPUs, eschewing the need for expensive GPUs. We train for 50 epochs, with a batch size of 512, using the AdamW optimizer \citep{kingma2015adam} with default learning rate of $0.001$ and weight decay of $0.01$. We do not use any other form of regularization (e.g., dropout). We select models based on validation loss with respect to a small number ($n=5$) of held out (language, trajectory) pairs.

As mentioned in \autoref{subsec:lilac-learning}, we assume a dataset of utterances and corresponding trajectories. These utterances consist of both the high-level task utterances (e.g., ``pick up the book and insert it into the bookshelf'') and correction utterances (e.g., ``tilt down a little bit''). We run the GPT-3 alpha labeling procedure as a preprocessing step, marking each example with the degree of context-dependence required, then train $\mathcal{F}_\theta$ jointly, on all of our data.

\begin{figure}[b!]
    \centering
    \includegraphics[width=\columnwidth]{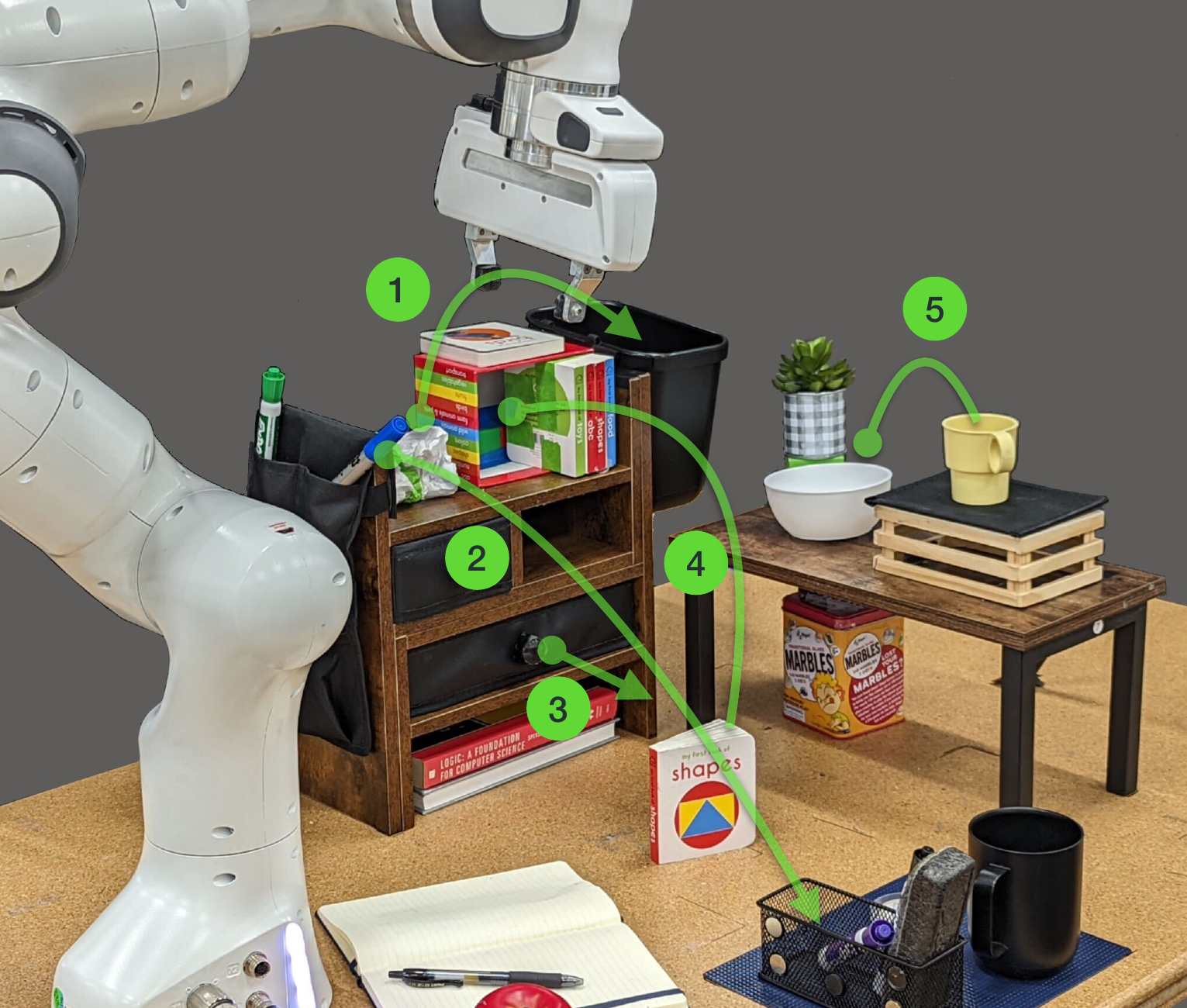}
    \vspace*{-7mm}
    \caption{Setup of our tabletop manipulation environment with sketches of our high-level tasks (further details in \autoref{sec:user-study-preliminaries}).}
    \label{fig:tasks}
\end{figure}

\section{User Study Preliminaries}
\label{sec:user-study-preliminaries}
\begin{figure*}[t!]
    \centering
    \includegraphics[width=\linewidth]{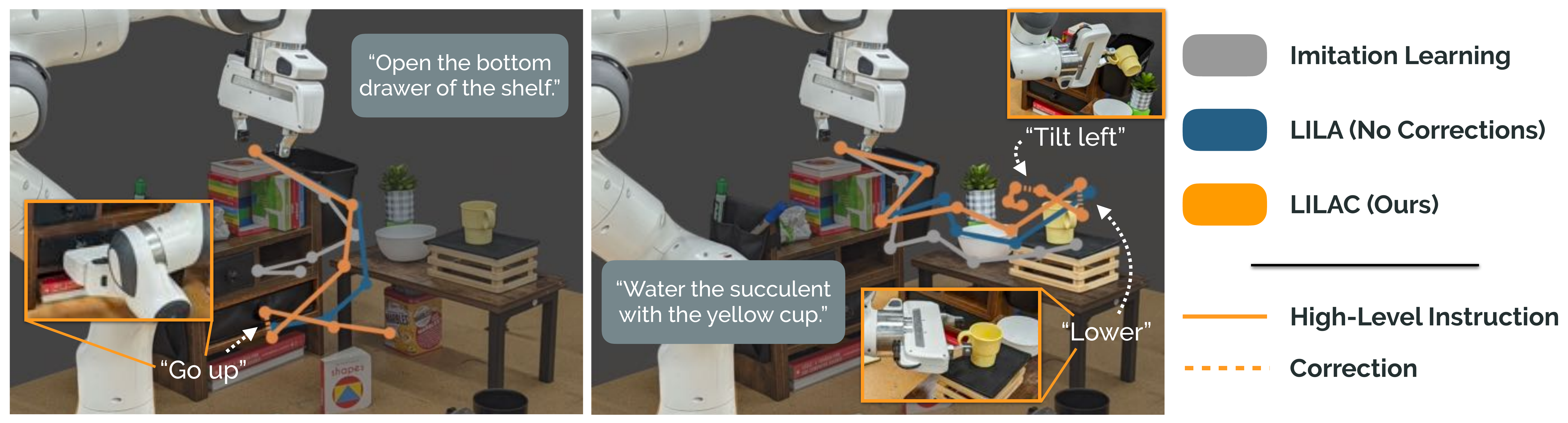}
    \vspace*{-7mm}
    \caption{Qualitative trajectories across the different control strategies for the \texttt{open-drawer} and \texttt{water-plant} tasks. The fully autonomous imitation learning approach fails to make it beyond the first stage of the task, while LILA is able to reach the drawer as well as the cup but fails to precisely aim and grasp the object. LILAC gets stuck at the same place, but is able to recover as the user issues low-level corrections to precisely maneuver the end-effector and fully complete the tasks.}
    \label{fig:overlays}
\end{figure*}

To evaluate LILAC with respect to prior methods for language-informed policy learning, we conduct a \textit{within-subjects} user study with $n = 12$ participants, with each participant evaluating LILAC against language-conditioned approaches for full and shared autonomy -- namely, a language-informed imitation learning baseline (``Imitation'') trained on the same demonstrations as LILAC, as well as a non-corrective shared autonomy baseline (``LILA''), also trained on the same demonstrations. The following sections detail the environment, tasks, data collection process, as well as user study procedure. Finally, we list our independent variables, dependent measures and concrete hypotheses.

\medskip

\noindent \textbf{Environment \& Tasks.} We consider a multi-task ``desk'' environment (\autoref{fig:tasks}) with the following tasks listed by complexity:
\begin{enumerate}
    \item \textbf{\texttt{clean-trash}}: throw away a piece of crumpled paper (deformable) into the black trash bin.
    \item \textbf{\texttt{transfer-pen}}: transfer the blue marker (upper left of \autoref{fig:tasks}) from the shelf into the metal tin holder (lower left).
    \item \textbf{\texttt{open-drawer}}: Open the bottom drawer on the shelf by grasping the small knob, and sliding out horizontally (requires fine-grained end-effector orientation control).
    \item \textbf{\texttt{insert-book}}: Pick up the book on the table by its spine, and insert it into the bookshelf (has only a few millimeters of clearance on either side).
    \item \textbf{\texttt{water-plant}}: Water the succulent (white bowl on the upper right of \autoref{fig:tasks}) using the water in the yellow cup (rather than actual water, we use marbles for easy cleanup).
\end{enumerate}
Each of the 5 tasks we define are difficult from a manipulation perspective, especially in the small data regime we operate in. For fine-grained comparison between the three approaches under study, we define a set of \textit{subtasks} to use to measure partial task success (turning a sparse full-task success rate into a denser measure of progress): a) \textit{reaching} the desired object to manipulate (e.g., the book in the \textbf{\texttt{insert-book}} task), b) successfully \textit{grasping} the manipulable object, c) \textit{transferring} the desired object to the target location (e.g., moving the water cup above the plant for the \textbf{\texttt{water-plant}} task), and finally, d) \textit{completing} the full task.

\medskip

\noindent \textbf{Demonstration \& Correction Data Collection.} For each task, we collected dense, human-guided kinesthetic demonstrations -- 50 full-task demonstrations total (orders of magnitude fewer demonstrations than what is typically required for fully autonomous instruction following approaches \cite{lynch2020grounding, mees2022matters}). The robot's proprioceptive state is encoded as the concatenation of its joint states (7-DoF; in radians), as well as the end-effector poses computed via forward kinematics (expressed as 3-DoF Cartesian position, and 3-DoF Euler angle orientation), and compute the high-level robot actions as the deltas in end-effector space between consecutive time steps in our demonstrations (resulting in 6-DoF high-level actions). We record our demonstrations and run our controllers at 10 Hz; we use Polymetis \citep{polymetis2021} as the basis for our robot control platform.

For LILAC, we additionally collect a small set of correction demonstrations (collected under 2 hours of interaction with the robot) with associated correction language utterances, spanning two loose categories: 1) directional corrections such as ``tilt down,'' ``to the right,'' ``rotate counterclockwise'', and 2) contextual referential corrections such as ``towards the blue marker'' or ``no, move down towards the knob on the drawer.'' We collect these demonstrations by replaying the full-task demonstrations, and sampling random intermediate states during playback to initiate corrections. The authors of this work served as the expert demonstrators for both full-task and correction demonstrations.

\medskip

\noindent \textbf{Participants \& Procedures.} All user studies were conducted subject to a university-approved IRB protocol, with participants recruited from a pool of 12 university students (8 male/4 female, age range 20-30 with mean 24.8). Of the 12 total participants, only 4 users had prior experience teleoperating a robot. All our studies used a Franka Emika Panda robot (as depicted in \autoref{fig:tasks}), a 7-DoF fixed-arm manipulator with a parallel-jaw gripper. In all settings, the maximum joint velocity norm of the Panda was bounded to 1 rad/sec, with conservative torque limits of 40.0 Nm. 

We conducted a \textit{within subjects} user study where each participant used all three candidate methods (denoted as \textit{Imitation}, \textit{LILA}, and \textit{LILAC (Ours)} in \autoref{fig:result-plots}) to complete 3 of the 5 high-level tasks. We shuffled the order of candidate methods between users to ensure a fair comparison. Upon starting, each user read a detailed written description of each control method (the explicit text can be found on the project website), viewed a video depicting the high-level task to perform, and were allowed a single ``practice'' session to with the control method in question. Each user performed two trials for a given task; we recorded partial success rates and asked the user to fill out a qualitative survey before switching to the next strategy.

\medskip

\noindent \textbf{Hypotheses.} In this study, we vary the control strategy (Imitation, LILA and LILAC) and use the objective measures of subtask and full-goal success rates to assess efficacy. We additionally track qualitative aspects such as ``ease of use,'' ``smooth control,'' and ``likelihood of using this control strategy again'' (the full set of qualitative measures can be found in \autoref{fig:result-plots}) by surveying our users via a 7-point Likert scale. We test the following two hypotheses regarding LILAC's performance relative to the baseline strategies: 

\smallskip

\textbf{H1 --} LILAC allows users to obtain higher subtask and full-goal success rates when completing complex manipulation tasks relative to specifying tasks for a language-conditioned imitation learning agent, or LILA trained on the same amount of data.

\smallskip

\textbf{H2 --} LILAC is qualitatively preferred by users over both baseline strategies in terms of overall usability measured by their subjective responses to the survey questions.

\medskip

\noindent \textbf{Baseline Implementations.} Both the LILA and Imitation (language-conditioned behavioral cloning) baselines are trained on the \textit{exact same data} as LILAC following nearly identical processes.\footnote{Implementations: \url{https://github.com/Stanford-ILIAD/lilac/tree/main/models}.} LILA is trained following the exact same architecture as LILAC from \hbox{\autoref{fig:system}}, \textit{without} the GPT-3 gating component. The imitation learning model is implemented as a \textit{history-aware} policy that is able to attend to prior states (in this work, we truncate history at 1 second). We use the same state and language encoder as LILAC, but use a 2-layer Transformer \hbox{\citep{vaswani2017attention}} to encode the full history sequence. We use again use a FiLM layer to fuse language and state embeddings, then use a final 2-layer MLP to directly predict the action to execute in the environment (for open-loop control).

\section{User Study Results}
\label{sec:results}
\begin{figure*}[t!]
    \centering
    \includegraphics[width=\linewidth]{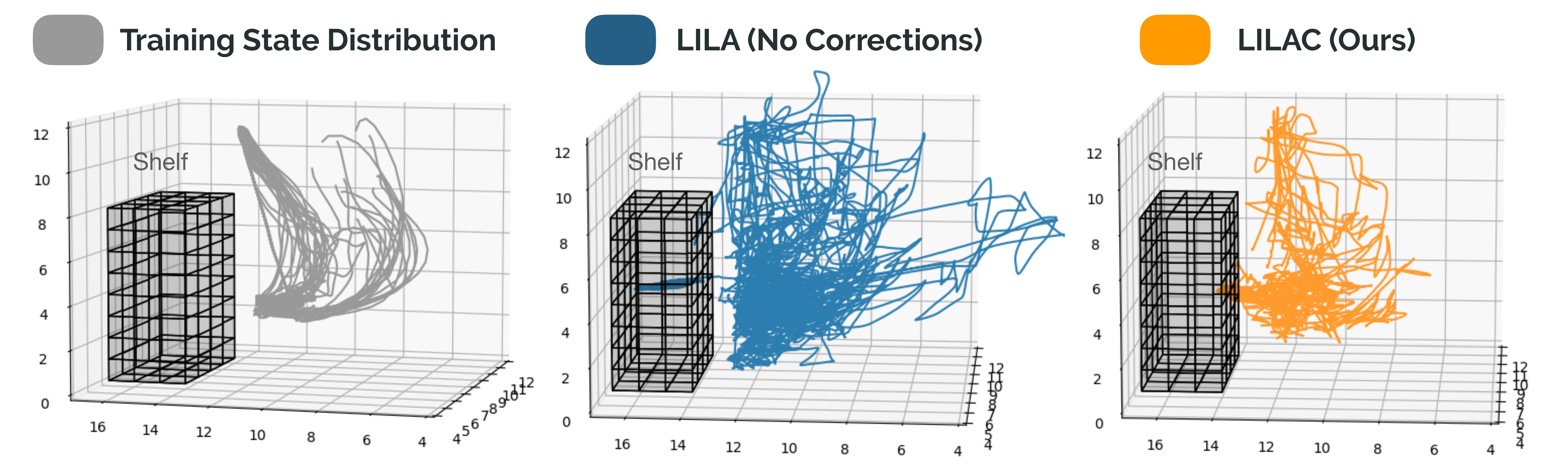}
    \vspace*{-7mm}
    \caption{Observed trajectories for LILA and LILAC on the \texttt{open-drawer} task (with train trajectories shown on the left). While LILA deviates from the observed state distribution, states traversed with LILAC are close to those seen at training.}
    \label{fig:traj-distribution}
\end{figure*}

We report the success rate for each subtask averaged across users in \autoref{fig:result-plots} (Left). Objectively, we find that LILAC achieves the highest success rate across all subtasks, and is significantly ($p < 0.05$) more performant than the imitation learning and LILA baselines for the latter three subtasks -- \textit{grasping}, \textit{transfer}, and \textit{full task completion} -- results that fully support \textbf{H1}. We also find that LILAC is subjectively preferred by users, as evidenced by \autoref{fig:result-plots} (Right). Looking at the survey results, we find that LILAC is significantly ($p < 0.05$) preferred on 6 out of the 7 metrics, including ``ease of use,'' ``intuitiveness,'' and ``willingness to use again,'' amongst others. The subjective results support \textbf{H2}; LILAC is favored for its \textit{adaptability}, allowing users to execute targeted, precise motions. 

\noindent \textbf{Visualizations.} To further understand the value of LILAC and incorporating online corrections, we visualize example trajectories for each of the three control strategies for two high-level tasks in \autoref{fig:overlays}. On the left are trajectories for the simple \texttt{open-drawer} task: we see that the fully autonomous imitation learning model fails to reach the drawer entirely, whereas LILA and LILAC are able to successfully reach the drawer, but get stuck trying to precisely aim and grasp the small knob. While LILA cannot recover, LILAC is receptive to the user's correction, producing a refined control space allowing for the user to complete the grasp and finish out the task. We see a similar story with the more difficult \texttt{water-plant} task: imitation learning fails catastrophically by knocking over the cup, causing irreversible damage to the environment, while LILA reaches the cup, but does not afford the user enough precision to make a successful grasp. With LILAC, two tightly sequenced corrections allow the user precise, targeted control, first in acquiring the cup, then in aligning the end-effector orientation to complete the pouring motion successfully. These trajectory visualizations offer insight into \textit{where} and \textit{when} corrections are most useful -- specifically showing the need for adaptivity in critical states.

\autoref{fig:traj-distribution} additionally plots the 3D end-effector trajectories (position; orientation is omitted for clarity) across all users for the \texttt{open-drawer} task for both LILA and LILAC, in addition to the trajectories represented in the training data. We find that states LILAC allows the users to stay closer to the training state distribution compared to LILA, further explaining LILAC's strong performance.

\section{Discussion}
\label{sec:discussion}
While the results of the user study are compelling, we find it important to be transparent about the shortcomings of the current approach, addressing possible avenues for future work.

\medskip

\noindent \textbf{Limitations.} Future work should address different types of language corrections that are more context sensitive; as a concrete example, due to the way we encode incoming corrections (as described in \autoref{sec:lilac}), we interpret each utterance on its own, independent of what was said previously. This is limiting for interpreting phenomena such as anaphora or implicit coreference -- corrections such as ``no, the other way,'' or ``undo that'' cannot be correctly interpreted using the current instantiation of the framework. Furthermore, we find that while corrections offer additional flexibility over the base shared autonomy control space, they can be easily overused -- we noticed that certain participants in our study quickly departed from the control space induced by the high-level instruction, instead opting to complete the bulk of the task in correction mode, effectively turning LILAC into a glorified end-effector control, with users moving one axis at a time. Work on making the underlying high-level control spaces more natural and intuitive -- \textit{naturalizing} the control interface -- will be crucial for scaling LILAC to more complex, temporally extended tasks where low-level corrections alone may not afford users enough expressivity to solve tasks. Some scenarios where this may occur would be in tasks requiring modulating 3+ degrees-of-freedom simultaneously, or where sequences of low-level corrections only allow users to make frustratingly slow progress at the task at hand (e.g., for long-horizon tasks like making a cup of tea). Finally, we find that corrections such as ``rotate'' or ``tilt'' can be ambiguously interpreted, with some users intending for the correction to be interpreted subject to their reference frame rather than the robot's reference frame, or vice-versa.

\medskip 

\noindent \textbf{Conclusion.} Throughout this work, we have argued that scalable systems for language-driven human-robot interaction \textit{must} be able to exhibit both \textit{adaptivity} and \textit{sample efficiency}. We identified the ability to handle \textit{online natural language corrections} as a way to enrich existing systems with such adaptivity, presenting LILAC -- Language-Informed Latent Actions with Corrections -- as a potential answer. LILAC is built within the shared autonomy paradigm whereby natural language utterances are mapped to meaningful, low-dimensional control spaces that humans can use to guide the robot, with each correction provided by the user working to \textit{refine} the underlying control space, allowing for precise, targeted control. Our user study comparing LILAC with language-conditioned imitation learning and language-informed shared autonomy shows the importance of being able to adapt to online corrections, as LILAC is both subjectively preferred by users and objectively performant than both baselines. LILAC marks a strong step forward in adaptive language-driven approaches for shared autonomy, and we hope that its core tenets of reliability, precision, and ease of use are carried forward throughout future work.

\begin{acks}
Toyota Research Institute (``TRI'') provided funds to support this work. This project was additionally supported by the Office of Naval Research, as well as by NSF Awards 2006388 and 2132847. Siddharth Karamcheti is grateful to be supported by the Open Philanthropy Project AI Fellowship. Finally, we would like to thank our anonymous reviewers.
\end{acks}

\clearpage
\bibliographystyle{ACM-Reference-Format}
\bibliography{refdb}

\end{document}